\documentclass{article}
\usepackage{comment}
\usepackage[paperheight=11.5in, paperwidth=9.5in]{geometry}
\usepackage{lscape}
\usepackage[utf8]{inputenc}
\usepackage[final]{pdfpages}
\usepackage{amsmath}
\usepackage{slashbox }
\usepackage[libertine]{newtxmath}
\usepackage{color}
\usepackage{setspace}
\usepackage{pdflscape}
\usepackage{graphicx} 
\usepackage{float}
\usepackage{algorithm}
\usepackage{algorithmicx}
\usepackage{algpseudocode}

\title{ Applications of Nature-Inspired Algorithms for Dimension {R}eduction: Enabling  Efficient Data Analytics}
\author{Farid Ghareh Mohammadi$^1$,  M. Hadi Amini$^2$,  and Hamid R. Arabnia$^1$\\1:Department of Computer Science,  Franklin College of Arts and Sciences, \\ University of Georgia,  Athens,  Georgia,  30601  \\
2: School of Computing and Information Sciences,  \\College of Engineering and Computing,  \\Florida International University,  Miami,  FL 33199 \\
Emails: farid.ghm@uga.edu,    amini@cs.fiu.edu,  hra@cs.uga.edu}
\date{}

\begin{document}

\maketitle
\section*{Abstract}

 In   \cite{ch1_farid},  we have explored the theoretical aspects of feature selection and evolutionary algorithms. In this chapter,  we focus on optimization algorithms for enhancing data analytic process,  i.e.,  we propose to explore applications of nature-inspired algorithms in data science. Feature selection optimization is a hybrid approach leveraging feature selection techniques and evolutionary algorithms process to optimize the selected features. Prior works solve this problem iteratively to converge to an optimal feature subset. Feature selection optimization is a non-specific domain approach.
 
 Data scientists mainly attempt to find an advanced way to analyze data n with high computational efficiency and low time complexity,  leading to efficient data analytics. Thus,  by increasing generated/measured/sensed data from various sources,  analysis,  manipulation and illustration of data grow exponentially. Due to the large scale data sets,  Curse of dimensionality (CoD) is one of the NP-hard problems in data science. Hence,  several efforts have been focused on leveraging evolutionary algorithms (EAs) to address the complex issues in large scale data analytics problems. Dimension reduction,  together with EAs,  lends itself to solve CoD and solve complex problems,  in terms of time complexity,  efficiently. In this chapter,  we first provide a brief overview of previous studies that focused on solving CoD using feature extraction optimization process. We then discuss practical examples of research studies are successfully tackled some application domains,  such as image processing,  sentiment analysis,  network traffics / anomalies analysis,  credit score analysis and other benchmark functions/data sets analysis.

 \textbf{Keywords:}  Dimension Reduction,  Data Science,  Hybrid Optimization,  Curse of Dimensionality (CoD),  Supervised Learning,  Unsupervised Learning,  Wrapper Feature Selection,  Classification,  Evolutionary Computation,  Swarm Intelligence,  Filter Feature Selection.

\section{Introduction}
\subsection{Overview}
Feature selection and evolutionary algorithms have been explored in \cite{ch1_farid}. In this chapter,   we aim to focus on an optimization approach for enhancing data analytic process. Feature selection optimization is hybrid approach leveraging pure feature selection techniques and evolutionary algorithms process to optimize the selected features. Researchers try to iterate this process until to converge to an optimal feature subsets. Feature selection optimization is non-specific domain approach which enable scientists to apply this to their data technically.
 
 Data scientists always attempt to find an advanced way to work with data that are successfully conducted in a short time with high computational efficiency and low time  complexity,  leading to efficient data analytics. Thus,  by increasing generated/measured/sensed data from various sources,  analysis,  manipulation and illustration of data grow exponentially. Due to the large scale data sets,  Curse of dimensionality (CoD) is one of NP-hard problems in data science. Hence,  several efforts have been focused on leveraging evolutionary algorithms (EAs) to address the complexity issues in large scale data analytics problems. Dimension reduction,  together with EAs,  lend itself to solve CoD and solve  complex problems,  in terms of time complexity,  efficiently. In this chapter,  we first provide a brief overview  of previous studies that focused on solving CoD using feature extraction optimization process. We then discuss practical examples of research studies are successfully tackled some application domains,  such as: image processing,  sentiment analysis,  network traffics / anomalies analysis,  credit score analysis and other benchmark functions / data sets analysis . 

In  \cite{ch1_farid},  we have comprehensively explored various evolutionary algorithms for data science,  including artificial bee colony (ABC),  ant colony optimization (ACO), Coyote Optimization Algorithm (COA),  genetic algorithm (GA),  grey wolf optimizer (GWO) and particle swarm optimization. These algorithms have real-world applications in the context of dimension reduction  \cite{mohammadi2014ifab, ahmed2017novel, hafez2015innovative,  liu2018hybrid,  kozodoi2019multi,  xue2018self,  harfouchi2018modified,  wang2018improved,  cao2018improved},  interdependent smart city infrastructures  \cite{amini2018panorama, amini2019sustainable2019},  optimization  \cite{gupta2018opposition,  li2002optimizing,  GA2001app}. This chapter focuses on  application of aforementioned evolutionary algorithms in dimension reductions  \cite{pathak2019GWO,  al2019binary,  shi2017efficient,  shi2018efficient,  chhikara2016hybridPSO, adeli2018imagePSO, rostami2016PSO, dong2017efficient, bui2019whale, tubishat2019improved}. As mentioned in  \cite{ch1_farid},   some studies deployed GA,  PSO,  or their combination as  effective tools for solving large-scale optimization problems,  including optimal allocation of electric vehicle charging station and distributed renewable resource in power distribution networks  \cite{PSO2017app, mozafar2017simultaneous},  resource optimization in construction projects  \cite{GA2001app},  and allocation of electric vehicle parking lots in smart grids  \cite{GA2014app}. Moreover,  we discuss some decent optimization using evolutionary algorithms including butterfly optimization algorithm (BOA),  chicken swarm optimization (CSO),  coral reefs optimization (CRO) and whale optimization algorithm (WOA).

Table \ref{tab:recentStudies} presents a representative information about the feature selection techniques using evolutionary algorithms. 

\begin{landscape}

\begin{table}[H]

    \centering
    \vline
    \begin{tabular}{c|l|l|l|l|l|l|}
     \hline
    Paper & Evolutionary algorithm & Proposed method &Feature Extraction type  & problem&domain & year \\
     \hline
       \cite{mohammadi2014ifab} & Artificial bee colony & IFAB &   FS (Wrapper)&Supervised & Image processing & 2014  \\
          \cite{mohammadi2014IFABKNN} & Artificial bee colony & IFAB-KNN &  FS (Wrapper) &Supervised & Image processing & 2014  \\ 
          \cite{mohammadi2017RISAB} & Artificial bee colony & RISAB &  FS (Wrapper) & Supervised & Image processing & 2017  \\
         
           \cite{xie2019unsupervised} & Artificial bee colony & ISD–ABC &  FS (Wrapper) & Unsupervised & Image processing & 2019  \\
          
            \cite{rao2019feature} & Artificial bee colony & ABCoDT  &  FS (Wrapper) & Unsupervised & Benchmark analysis& 2019  \\
           
             \cite{zhang2019cost} & Artificial bee colony & TMABC-FS  &  FS (Wrapper) & Unsupervised & Benchmark analysis& 2019  \\
            
          \cite{peng2018improved}& Ant colony optimization & FACO & FS (Wrapper) & Supervised & Network anomalies analysis &2018\\
       \cite{ke2008efficient}& Ant colony optimization & ACOAR & FS (Filter) &Supervised& Benchmark analysis&2008\\
           \cite{tabakhi2014unsupervised}& Ant colony optimization & UFSACO  & FS (Filter) & Unsupervised & Benchmark analysis &2014\\ 
           \cite{moradi2015integration}& Ant colony optimization & graph-ACO  & FS (Filter) & supervised & Benchmark analysis&2015\\
            
         \cite{kabir2012new}& Ant colony optimization & ACOFS  & FS (Filter and wrapper) & Supervised & Benchmark analysis&2012\\
        
           \cite{arora2019binary} & butterfly optimization algorithm  & bBOA  & FS (wrapper)& Supervised &Benchmark analysis& 2019\\
          
          \cite{yan2019hybrid} & Coral Reefs Optimization & BCROSAT  & FS (wrapper)& Supervised &Biomedical analysis& 2019\\
         
       \cite{ahmed2017novel} & Chicken swarm optimization & CCSO  & FS (wrapper)& Supervised &Benchmark analysis& 2017\\
       \cite{hafez2015innovative} & Chicken swarm optimization & CSO-KNN & FS (wrapper)& Supervised &Benchmark analysis& 2015\\
       \cite{liu2018hybrid} & Genetic Algorithms & HGAWE & FS (wrapper and embedded)& Supervised & Biological analysis& 2019\\
        \cite{kozodoi2019multi} & Genetic Algorithms & NSGA-II & FS (wrapper)& Supervised & Credit scoring analysis& 2019\\
       
       \cite{pathak2019GWO} & Grey Wolf Optimizer&LFGWO & FS (wrapper)&Supervised & Image processing&2019 \\
       \cite{al2019binary} & Grey Wolf Optimizer and PSO &LFGWO & FS (Wrapper) &Supervised & Benchmark analysis&2019 \\
      
      \cite{al2019binary} & Fish swarm algorithm  & FSANRSR & FS (Wrapper) &Supervised & Rough set-Benchmark analysis&2019 \\
         
       \cite{shi2017efficient} & - &  PCA &Dimension reduction&Supervised&  Network Traffics analysis& 2017\\
       \cite{shi2018efficient} & - & PCA &Dimension reduction &Supervised&   Network Traffics analysis& 2018\\
          \cite{chhikara2016hybridPSO} & Particle swarm optimization & HYBRID PSO & FS (Filter and wrapper) &  Supervised & Image processing & 2016\\
          
           \cite{adeli2018imagePSO} & Particle swarm optimization & APSO &  FS (Filter ) &  Supervised &Image analysis & 2016\\
              \cite{kaur2018feature} & Particle swarm optimization & MI-APSO &  FS (Filter ) &  Supervised &Image analysis & 2018\\
        \cite{dong2017efficient} & RelieF  and PSO  & RFPSO & FS ( Filter and
       wrapper)&Supervised & Network Traffics analysis&2017   \\
       
     \cite{bui2019whale} &  Whale optimization algorithm  & WANFIS & FS (        wrapper)&Supervised & Image processing&2019   \\
        \cite{tubishat2019improved} &  Whale optimization algorithm  & IWOA & FS (        wrapper)&Supervised & Sentiment analysis &2019   \\
      
       \hline
        
    \end{tabular}
  
    \caption{An overview of previous studies on supervised/unsupervised data sets applying hybrid (combined) feature selection.}
    \label{tab:recentStudies}
\end{table}

\end{landscape}

 \subsection{Organization}

\begin{figure}[H]
    \centering
    \includegraphics[height=3.9in]{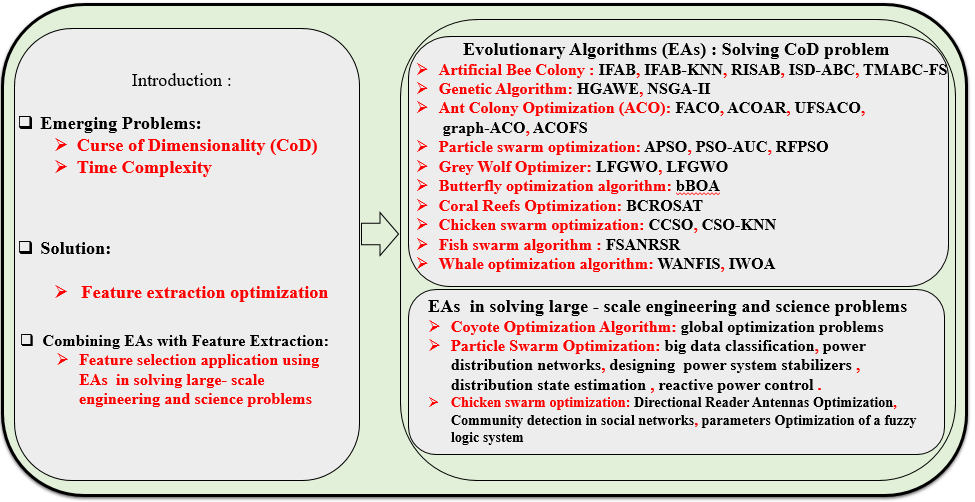}
    \caption{Overall structure of this study}
    \label{fig:General_Structure}
\end{figure}

The rest of this study is organized as follows. In section 2,  we have discussed the applied evolutionary algorithms and their application. Then,  we introduce hybrid feature selection methods using evolutionary algorithms and their application in solving engineering and science problems. In general,  Figure \ref{fig:General_Structure} represents the overall structure of this study.

\section{Application of evolutionary algorithms}
Engineering,  industries,  scientists consider EA at the very final plan. They try to solve their problems easily in low time complexity run time. They used plenty of algorithms to find an optimal solution for their problems,  but all of them failed. Some problems they are struggling with are NP-hard problems which are required to ponder the problems deeply. Therefore,  final plan for researchers which is left is adopting evolutionary algorithms. Therefore,  research scientist do not use EAs for solving simple problems and only consider them for challenging issues and NP-hard problems.  It means that EAs have wide variety of applications and are not limited to a specific problems. In this section we will review some applications of them in Image processing  \cite{mohammadi2014ifab,  elaziz2018galaxies,  xie2019unsupervised},  optimization problems \cite{wang2018improved}. Thus,   Evolutionary algorithms are an impactful methods to address different NP-hard problems \cite{gong2016discrete}.Table \ref{tab:recentStudies} presents a representative information about the hybrid (combined) feature selection techniques using evolutionary algorithms on supervised / unsupervised data sets. Based on  \cite{liu2018hybrid,  rostami2016PSO},   hybrid (combined) methods are newly introduced,  which mixes evolutionary algorithms together with filter based or wrapper based algorithms. So,  All proposed methods you see in the table are hybrid methods. Table \ref{tab:AbbreTable} provides complete definitions for abbreviation which are used in this chapter.

\subsection{Feature extraction optimization }
\begin{figure}[H]
    \centering
    \includegraphics[height=3.5in]{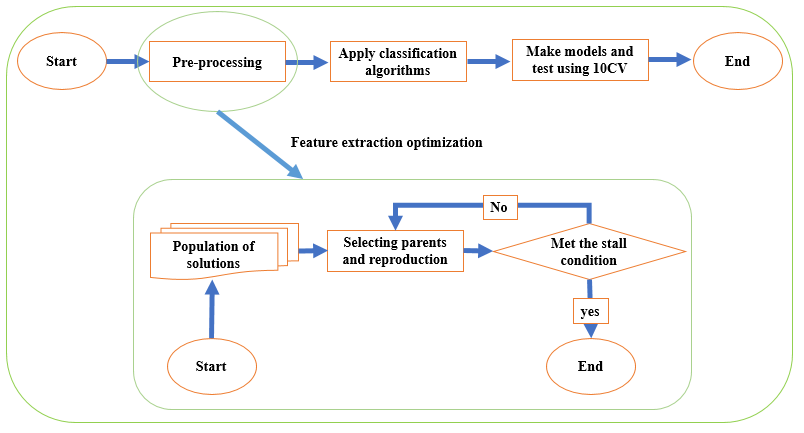}
      \caption{Evolutionary algorithms process for  feature extraction}

    \label{fig:EV_FE-general}
\end{figure}

Finding a proper subset of features or generating a new sets of features while decreasing the dimension of data sets and improving the performance is still a NP-hard  \cite{ermon2013taming,  kouiroukidis2011effects}  problem and scientists try to solve this problem. FE has been successfully done in several domains  \cite{shi2018efficient,  vanaja2018novel,  hancer2018pareto} and it is not domain specific.  Here,  we dig into some specific important feature selection algorithms like IFAB \cite{mohammadi2014ifab}. IFAB is a feature selection method has applied on digital images. Figure\ref{fig:EV_FE-general} depicts a procedure of FE approach using EAs in an abstract way. It shows that EAs are adopted in pre-processing section to help scientists to reduce the number of feature properly. In general,  we try to use a classification method to learn from train data and make a model. Then,  leverage the generated model to predict un-labeled test data and calculate the performance of the classifier. As we discussed,  classifiers fail to learn a large amount of data due to the CoD problem,  So we discuss feature extraction optimization,  how it helps classifier to learn the train data without struggling with over-fitting or under-fitting problems.
\subsubsection{Feature selection for image classification }
Due to growth of generating images in different areas and networks,  transferring,  distributing images,  image labeling and classification raised the majority of researchers' attention. To do that,  scientists have implemented plenty of tools and packages and libraries,  such as deep learning  \cite{khan2019novel},  generative adversarial networks (GANs)  \cite{roy2018semantic},  convolutional neural network (CNN) \cite{zhang2019sar},  recurrent neural network (RNN)  \cite{hang2019cascaded}. These provided algorithms have been successfully applied and determined very low loss in their accuracy. However,  they still suffer from handling dig data,  CoD problem and time complexity. Thus,  researchers attempted to take advantage of leveraging evolutionary algorithms to address these problems properly. Image classification is one of  Image stegnalysis has been quite an emerging challenges in large-scale engineering and science,   particularly in advanced image processing. The importance of issue has proliferated significantly last 20 years after the tragedy happened in 2001.  \\[2pt]
\setlength{\parindent}{1cm} 
\begin{table}[H]

    \centering
    \vline
    \begin{tabular}{c|l|}
     \hline
    Abb & Definition\\
     \hline
     ABC & Artificial bee colony \\
     ACOAR & Ant colony optimization attribute reduction \\
      BSO & Bat swarm algorithm \\   
      BCO & Bee colony optimization \\
      BOA & Butterfly optimization algorithm \\
      COA & Coyote Optimization Algorithm \\
      CoD & Curse of dimensionality\\
      CSO & Chicken swarm optimization\\
      CCSO & chaotic chicken swarm optimization \\
      CRO & Coral reefs optimization\\
      DA & Dragonfly algorithm \\
      EAs & Evolutionary algorithms\\
      FS   & Feature selection\\
      FSA & Fish swarm algorithm\\
      GA & Genetic algorithm \\ 
      GWO & Grey Wolf Optimizer \\
      IFAB &   Image steganalysis using FS based on ABC \\
      IoT & Internet of things \\
      ILS & Iterated local search \\
      IWOA & Improved whale optimization algorithm \\
      PCA  & Principal component analysis \\
      PEAs & Parallel evolutionary algorithms\\
      RFPSO & RelieF  and PSO  algorithms  \\
      RL   & Representation learning\\
      RISAB & Region based Image Steganalysis using Artificial Bee colony \\
        SLS & Stochastic local search \\
      SSGA & Steady state genetic algorithm  \\
      SVD & Singular value dimension \\
      SVM & Support vector machine\\
      TMABC-FS & Two-archive multi-objective artificial bee colony algorithm for FS\\ 
     
      WOA & Whale Optimization Algorithm\\
      WANFIS &Whale adaptive neuro-fuzzy inference system\\
       \hline
        
    \end{tabular}
  
    \caption{Abbreviation of words}
    \label{tab:AbbreTable}
\end{table}
 \textbf{$\bullet$ Application of artificial bee colony in feature selection}: \\[0.2pt]

Image steganalysis using feature selection based on artificial bee colony (IFAB) is presented  by Ghareh Mohammadi and Saniee Abadeh in  \cite{mohammadi2014ifab}.  IFAB has an optimized feature extraction process and is categorized into wrapper-based feature selection methods. This method works properly for examining input digital images and distinguishing cover images from stego images. A stego image is a cover image with an embedded message. The goal of IFAB is to decrease time complexity of training model while improving the classifier's performance.

\begin{figure}[H]
    \centering
    \includegraphics[height=3.3in]{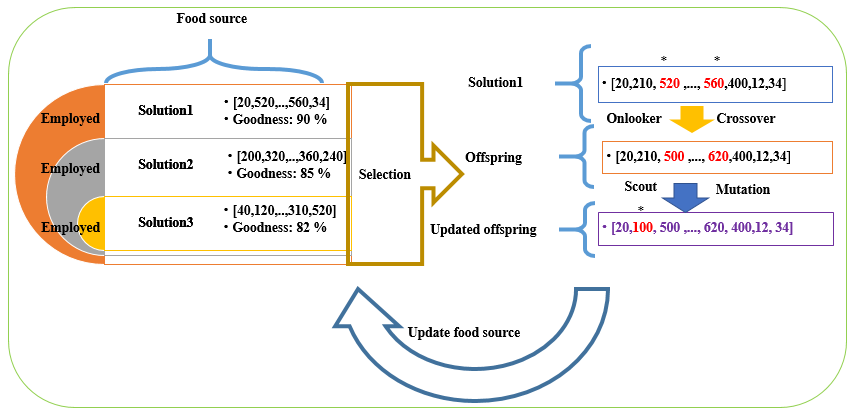}
  \caption{An overview of Image steganalysis using Feature selection based on Artificial Bee colony (IFAB)  \cite{mohammadi2014IFABKNN}}
   \label{fig:EV_FE_IFAB-general }
  \end{figure}

Figure \ref{fig:EV_FE_IFAB-general } presents IFAB's process in detail stating that how IFAB technically combine three types of bees to optimize the feature extraction process. First of all,  employed bees are applied to generate food sources and a goodness of each food source is generated using support vector machine(SVM). Fitness function leverage SVM to compute the goodness of each food source. According to the \ref{eq.fitness} fitness function is calculated. f[i] goes for each food source and $P_i$ stands for the accuracy of SVM per each food source. Each food source stands for a solution to the problem. A solution would present the number of features have selected to reduce the dimension of given data set. This step is done for each employed bee population. The employed bees and onlooker bees have the same population and equal to the dimension of input data. For instance,  SPAM data set has 686 features which is equal to the population of them. 
\begin{align}
   Fitness (f_i)=
\begin{cases}
    \frac {1}{1+P_i} & \text{if } P_i > 0\\
         1+abs(P_i) & \text{if } P_i <  0 \label{eq.fitness}
\end{cases}
\end{align}

Second step is choosing a food source based on (using equation \ref{eq.SelectFS} the goodness of the food source to exploit it by onlooker bees. Once offspring (new solution) is generated,  it is time for checking the condition,  limit. 

\begin{equation}
{V_i}= {f_i}+{v * (f_i-f_j)}. ,  v =[-1, 1]
 \label{eq.SelectFS}
\end{equation}

Where j is a random number between 1 and N. i, j stands for features index in a input data set. N is the upper bound of number of features. If the solution's performance stopped improving within the pre-defined iteration equal to the limit,  then,  in third step,  Scout bee is selected to explore  using equation \ref{eq.Scout} the area of the food source,  and  tries to update the solution by a pre-adjusted rate. $X_max$,  $X_min$ shows the upper bound and lover bound of population,  respectively.
\begin{equation}
{X_i}= {X_{max}}+{v'} * { ({X_{max}}-{X_{min}})} ,  v' =[0, 1] 
\label{eq.Scout}
\end{equation}

Each iteration the best food source with respect to the goodness value is reserved and ABC updates food sources by replacing the minimum food source(minimum value of goodness) with new food source if it's goodness outperforms minimum food source. ABC terminates by full-filling the condition or iteration is completed. Finally,  ABC returns the best food source during whole iterations. This food source involves a number of relevant and important features and yields a better result.

IFAB-KNN is an advanced ABC for image steganalysis to enhance IFAB performance proposed by Ghareh Mohammadi and Saniee Abadeh in  \cite{mohammadi2014IFABKNN}. IFAB-KNN stands for one of the wrapper-based feature selection algorithm. Fitness function takes advantage of a lazy algorithm,  k-Nearest Neighbor(KNN),  within ABC and enables ABC to examine each subset of features deeply. Keeping the same number of selected features,  IFAB-KNN outperforms IFAB with the tuned hyper-parameters.  Xie \emph{et al}  \cite{xie2019unsupervised} produced another unsupervised feature selection algorithm using ABC to classify hyperspectral images,  ISD–ABC.

\begin{figure}[H]
    \centering
    \includegraphics[height=3.4in]{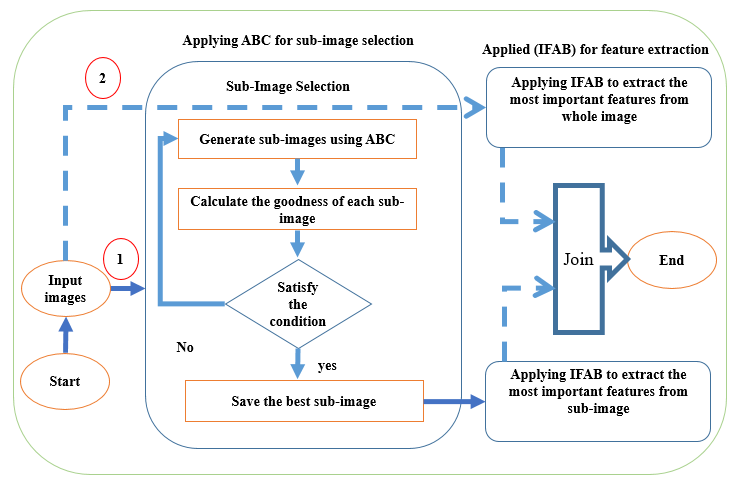}
    \caption{The process of feature selection using RISAB}
    \label{fig:RISAB}
\end{figure}

Ghareh mohammadi and Saniee Abadeh  \cite{mohammadi2017RISAB} presented another hybrid approach to feature selection,   combination of data and images. They proposed RISAB,  Region based Image Steganalysis using artificial bee colony by leveraging the IFAB characteristics. The goal of RISAB is to find the location of image that does not follow the harmony of the whole images. by finding special pixel or sets of pixels,  RISAB could distinguish stego images from cover images. In RISAB,  the researchers first applied ABC to find the most probable sub-image that carries the embedded data,  which would be messages,  images,  etc. ABC was tailored to focus on image spatial domain which was one of the important  challenging issues. After that,  there are one given input data and one sub-image of input data. Then,  researchers tried to apply IFAB on both of them. They were able to improve the performance of feature extractors like SPAM and CC-PEV,  even IFAB. Figure \ref{fig:RISAB} presents overall stepts of RISAB,  for more information you may read this paper  \cite{mohammadi2017RISAB}.

\setlength{\parindent}{1cm}  \textbf{$\bullet$ Application of particle swarm optimization in feature selection}: \\[0.2pt]

Chhikara \emph{et. al}  \cite{chhikara2016hybridPSO} proposed HYBRID, a new approach using PSO for solving CoD problem in image steganalysis. They proposed a combined filter and wrapper based feature selection approach to deal with high domensionality problem in image steganalysis. Authors tested HYBRID on data which had extracted using feature extraction methods,  using images which were attacked by steganography algorithms. The HYBRID enhanced the classification accuracy image steganalysis. Chikara and Kumari in  \cite{kumari2017GLBPSO} presented  a new wrapper-based feature selection for image steganalysis,  named Global Local PSO (GLBPSO) which leverages backpropagation
neural networks to evaluate the selected feature subsets. 

Adeli and Broumandnia in  \cite{adeli2018imagePSO}  introduced another filter-based feature selection algorithm for steganalysis named,  an Adaptive inertia weight-based PSO (APSO) where the inertia weight of PSO is adaptively adjusted leverage of three components,  such as: average distance of particles,  the swarm diameter,  and average velocity of particles. Rostami and Khiavi \cite{rostami2016PSO} take advantage of a novel fitness function which uses Area Under Curve (AUC) to evaluate feature subset. The Final accuracy of APSO yields a better result in comparison with IFAB,  however,  their time complexity problem still remains for a CoD problems.    

\setlength{\parindent}{1cm}  \textbf{$\bullet$ Application of grey wolf optimizer in feature selection}: \\[0.2pt]
Pathak \emph{et al}  \cite{pathak2019GWO} proposed a new feature selection algorithm, LFGWO which has been used to classify stego images from cover images. 
GWO has been widely used to solve large scale optimization,  engineering,  science problems,  such as global optimization tasks \cite{gupta2018opposition},   \cite{al2019binary} optimized feature selection algorithm using combining GWO and PSO called PSOGWO.
\subsection{Feature selection for network traffic classification}
Internet traffics growth has increased unexpectedly due to expanding new technologies and data comes from every where using internet of things (IoT). However,  as lack of traditional traffic classification approaches,  researchers  have been applying traffic classification using ML techniques. Current and simple ML techniques may not turn into a optimum solution because of very high dimensionality.

Having multi-class data sets which are imbalance become another emerging challenge for scientist. ML algorithms struggles with the data and  does not yield high recall for the minority classes. Researchers in their studies have proposed different hybrid approaches to address these problems. Scientists worked on data in pre-processing phase by using re-sampling approaches,  cost-sensitive approaches and feature extraction approaches which has very high impact on the classification process.
Dong \emph{et al} \cite{dong2017efficient} proposed a new hybrid method using PSO address the aforementioned problems. Researchers in this paper presented a new hybrid feature selection algorithm combining RelieF and PSO algorithms called RFPSO.
According to the figure ,  RFPSO has two main steps to follow. First step goes for initialization of RFPSO,  and second step is fitness function selection

Relief is an feature selection algorithm proposed by Kira and Rendell in 1992  \cite{kira1992practical} . It is one of the filter-based feature selection that works deeply on relationship among features. It was firstly presented to address binary classification problems including discrete or numerical features. 
RFPSO calls the fitness function which stands for the inconsistency rate defined in equation \ref{eq.RFPSO}  \cite{kira1992practical} . T stands for the total number of instances and the number of inconsistencies goes for N. They considered three popular criteria to evaluate their work. one Recall,  one Precision and one $F_{measure.}$

\begin{equation}
fitness= \frac {N}{T}. 
 \label{eq.RFPSO}
\end{equation}

Shi \emph{et al}  \cite{shi2017efficient,  shi2018efficient} aimed to classify traffic data by proposing a robust feature extraction and feature selection algorithms which leverage customized PCA. Hamamoto \emph{et al}  \cite{hamamoto2018network} used GA to detect network anomaly detection. GA is applied to create a digital signature of network segment by leveraging flow analysis. The information are provided and extracted from network flows which have been used to predict networks traffic action with in short time period. .They combined Ga with fuzzy logic to improve their result. Actually,  fuzzy logic Fuzzy enables the Ml algorithm decides whether a sample represents an anomaly or not.

\begin{figure}[H]
    \centering
    \includegraphics[height=3.4in]{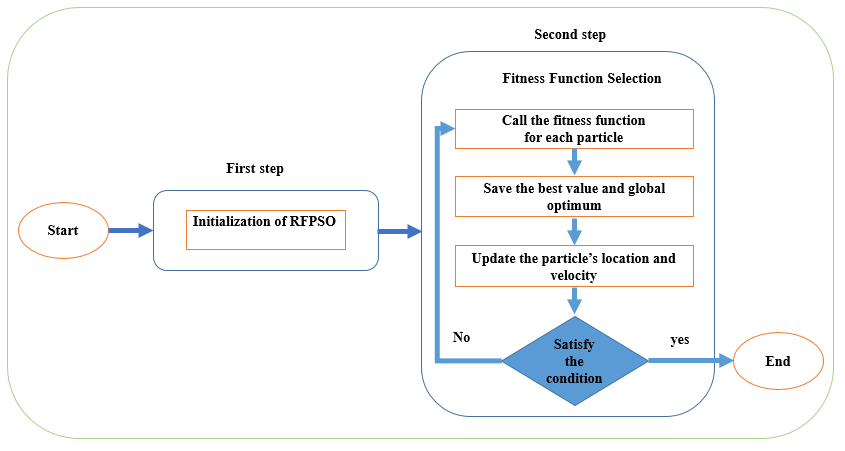}
    \caption{The process of feature selection using RelieF  and PSO  algorithms (RFPSO)}
    \label{fig:RFPSO}
\end{figure}

\subsection{ Feature selection benchmarks}
Ahmed \emph{et al}  \cite{ahmed2017novel} presented an advanced feature selection algorithm using chaotic chicken swarm optimization (CCSO). They used CCSO to cover the problem of stalling in local minimum which is one of the important and common problems of  traditional evolutionary and swarm algorithm. Unlike other evolutionary algorithms,  it remembers both minimum and maximum value of each solution to optimize the search step.  They used five different data sets,  such as spambase,  wbdc,  ionosphere,  lung and sonar. They compared their result with other evolutionary algorithms like PSO,  binary CSO,  bat swarm algorithm (BSO) and dragonfly algorithm (DA).

Ant colony optimization attribute reduction (ACOAR) is another feature selection algorithm which reduce the number features using filtering based algorithm.  ACOAR is introduced in  \cite{ke2008efficient} and leverages the ACO process to improve the performance of algorithm in rough set theory. Moreover,  Tabakhi \emph{et al}  \cite{tabakhi2014unsupervised} proposed another yet novel approach for feature selection algorithm for unsupervised data,  called UFSACO .  UFSACO looks for the optimal and the most relevant feature subset,  however,  it does not take advantage of learning algorithms to that end.Furthermore,  it goodness of features relation is computed with respect to the similarity among features.  UFSACO aims to the minimize the redundancy of features.  Moreover,  ACOFS,  a hybrid ant colony optimization algorithm presented in. \cite{kabir2012new},  graph-based feature selection using ACO  \cite{moradi2015integration}.

\setlength{\parindent}{1cm}  \textbf{$\bullet$ Recent advances in feature selection optimization}:\\
We have included and classified the most popular and important research studies with respect to the evolutionary algorithms. In this section,  we would like to add last,  but not the least,  yet cutting-edge studies attempted solve to the NP-hard problems, 
Rao \emph{et al}  \cite{rao2019feature} presented another hybrid feature selection method using bee colony and leveraging gradient boosting decision tree to address the NP-hard problems like curse of dimensionality. Bui \emph{et al}  \cite{bui2019whale} proposed a hybrid feature selection for land pattern classification,  using Whale Optimization Algorithm (WOA) and adaptive neuro-fuzzy inference concepts. Another improved WOA is presented in  \cite{tubishat2019improved} to Arabic sentiment analysis by feature selection. Kozodoi \emph{et al}  \cite{kozodoi2019multi}  levergaed profit measures to propose a wrapper feature selection algorithm  using NSGA-II genetic algorithm. Yan \emph{ et al}  \cite{yan2019hybrid} improved yet another evolutionary algorithm,  Coral Reefs Optimization (CRO),  to select the best matched feature subsets, called BCROSAT to apply on biomedical data. Arora and Anand    \cite{arora2019binary} introduced a binary variants of a new evolutionary algorithm,  the butterfly optimization algorithm (BOA),  to select the most important features. Zhang \emph{et al}  \cite{zhang2019cost} proposed new cost-sensitive multi-objective ABC-based feature selection,  TMABC-FS.  Pierezan and Coelho  \cite{pierezan2018coyote} proposed new optimization algorithm based on COA,  for solving global optimization problems.
\section{Discussion}
NP-hard problems always become the most challenging issues,  most frequently seen in engineering and science. In this chapter,  we address one of the emerging NP-hard problems in data science,  the curse of dimensionality (CoD). A large number of research studies have attempted to solve this problematic issue. Researchers continue to publish new papers on this problem since they have not found an algorithm that performs in an accurate and robust way. Researchers did not obtain a better result by only using basic machine learning algorithms, such as support vector machine (SVM) or K-nearest neighbour (KNN). During the last two decades,  scientists started to apply evolutionary algorithms to this issue and will continue publishing unlimited papers on this topic. Furthermore,  Data scientists have significantly increased the number of domains to which they could apply evolutionary algorithms. So,  lately,  the number of data has proliferated and lead scientist to find a new type of data called Big data. Thus,  working with data,  including manipulating data and finding the pattern governing whole data becomes harder. Not only do we have different kinds of supervised data set with ground truth,  but also this new type of data introduced a new concept,  the unsupervised data set.  Scientists proposed a technical way to deal with different groups of data,  meanwhile their approaches have to be able to learn unsupervised data sets,  too. The future work would be trying to propose a solution to apply evolutionary algorithms to help representation learning which enables scientists to find patterns and important features independent of the ground truth and enables classifiers to learn supervised and unsupervised data properly to avoid over-fitting.

\section{Conclusion}

Data science has some defects because of some data problems due to unexpected growing amount of it. The main problem of data is the curse of dimensionality (CoD),  which causes yet another important problem,  high time complexity. To solve CoD problem,  researchers have been proposed different feature selection optimization techniques by selecting the most relevant and optimal features,  tries to evaluate supervised/ unsupervised data sets. In this chapter,  we provided the practical examples of applied evolutionary algorithm for feature selection optimization and reviewed emerging optimization application. Furthermore,  scientists have adopted evolutionary algorithms  like IFAB and tailored them towards their goals. Moreover,  having applied evolutionary algorithms to select the most important features,  researchers improved the performance of classification algorithms. For instance,  IFAB decreased the dimension of a given data set intensely,  while also enhancing the performance of the support vector machine (SVM) significantly. In addition,  IFAB decreased the time to learn model,  and improved time complexity. In this chapter,  we categorized optimization method according to the evolutionary algorithms and their applications. It is noteworthy that evolutionary algorithms have been applied successfully on most of engineering,  science and even biology and medical domains which make them powerful and robust enough. This study provides the required information for the researchers who plan to pursue research on dimension reduction and large-scale optimization problems.

\bibliographystyle{unsrt}
\bibliography{bib.bib}
\end{document}